\documentclass[11pt,letterpaper]{article}

\usepackage[margin=1in]{geometry}
\usepackage[utf8]{inputenc}
\usepackage[T1]{fontenc}
\usepackage{amsmath,amssymb}
\usepackage{graphicx}
\usepackage{subcaption}
\usepackage{booktabs}
\usepackage{microtype}
\usepackage[round]{natbib}
\usepackage{url}
\usepackage{hyperref}
\hypersetup{colorlinks=true,linkcolor=blue,citecolor=blue,urlcolor=blue}
\usepackage{todonotes}

\graphicspath{{images/}}

\title{\bf Score Distributions, Not Cells: Evaluating\\Single-Cell Perturbations Under Class Overlap}

\author{
  \begin{tabular}[t]{@{}c@{}}
    \small\bfseries Youssef Marrakchi\textsuperscript{*,1}\\
    {\small\texttt{youss24@mit.edu}}
  \end{tabular}
  \hspace{1.2em}
  \begin{tabular}[t]{@{}c@{}}
    \small\bfseries Davide D'Ascenzo\textsuperscript{*,2,3}\\
    {\small\texttt{davide.dascenzo@unimi.it}}
  \end{tabular}
  \hspace{1.2em}
  \begin{tabular}[t]{@{}c@{}}
    \small\bfseries Sebastiano Cultrera di Montesano\textsuperscript{4}\\
    {\small\texttt{scultrer@broadinstitute.org}}
  \end{tabular}
}
\date{}

\begin{document}
\maketitle
{\renewcommand{\thefootnote}{}\footnotetext{%
\textsuperscript{*}Equal contribution.\quad
\textsuperscript{1}Department of Electrical Engineering and Computer Science, Massachusetts Institute of Technology, Cambridge, MA, USA.\quad
\textsuperscript{2}Department of Computer Science, University of Milan, Milan, Italy.
\quad
\textsuperscript{3}Department of Control and Computer Engineering, Politecnico di Torino, Torino, Italy.\quad
\textsuperscript{4}Eric and Wendy Schmidt Center, Broad Institute of MIT and Harvard, Cambridge, MA, USA.}}

\begin{abstract}

Most classification problems assume the classes are roughly separable, so that an individual sample can usually be assigned to one class. Single-cell perturbation data violates this assumption: two perturbations can produce different populations of cells while overlapping so much that an individual cell could belong to either. Per-cell accuracy then measures this overlap rather than model quality. We see this on Tahoe-100M and the Virtual Cell Challenge, where a linear classifier, an MLP, and a Transformer all plateau near macro-F1 0.2–0.3 even though almost every pair of perturbations is statistically distinguishable.

The fix is to score perturbations across the whole population rather than cell by cell. We average a classifier's per-cell probability vectors over all cells of a perturbation to form a population profile, then rank candidate perturbations by this profile; we call the resulting score the Classifier Discrimination Score (CDS). Taking the top-ranked class recovers the winning perturbation. It needs no retraining, costs linear time in the number of cells, and recovers near-perfect identification from the same weak models. CDS differs from the pseudobulk-based Perturbation Discrimination Score (PDS) used in recent benchmarks only in where the average is taken, raw gene expression for PDS versus a learned discriminative space for CDS, and identifies the true perturbation more reliably on both datasets, with the gap widening as cells grow scarce. Because a metric that misranks the ground truth will misrank the models scored against it, per-cell accuracy and raw-pseudobulk scores should be used with caution when comparing perturbation models.

\end{abstract}

\section{Introduction}
\label{sec:intro}

Perturbation experiments read out by single-cell RNA sequencing have become a cornerstone of modern biology. Rather than measuring one average response per condition, they return a whole distribution of responses, one transcriptomic profile per cell \citep{luecken2019current}. The same drug or genetic intervention, applied to a population of otherwise similar cells, produces a spread of outcomes. Some of that spread is biological, since cells differ in cell-cycle stage and recent history when the perturbation arrives \citep{elowitz2002stochastic}. Some of it is technical, since sequencing captures only a noisy sample of each cell's transcripts \citep{kharchenko2014bayesian}. Atlases like Tahoe-100M now hold hundreds of millions of these profiles across thousands of perturbations and cell lines \citep{zhang2025tahoe}.

These atlases have spurred a wave of computational models. Most tackle the forward problem: given control cells and the name of a perturbation, predict the distribution of cells it would produce \citep{lotfollahi2019scgen, lotfollahi2023predicting, roohani2024predicting, adduri2025predicting}. This paper is about the reverse. Given a population of cells, which perturbation produced them? The question matters on its own, for example in working out a drug's mechanism of action \citep{haber2025heimdall}. But it also sits inside every discrimination-based evaluation metric. To score a model's prediction, you ask whether its cells match the intended perturbation and not some other one \citep{wu2024perturbench, roohani2025virtual}. We study that question on real cells, where the true label is known. This lets us measure how well a score discriminates perturbations, without folding in the quality of any single prediction model, as in recent analyses of these metrics \citep{liu2025effects}.

The most natural way to approach this is by training a classifier. Show it labeled cells, let it learn to name the perturbation behind each one, and evaluate it by its cell-level accuracy \citep{wu2024perturbench, ahlmann2025deep}. The habit of judging classifiers by per-sample accuracy comes from problems where the classes are approximately separable: a picture is a cat or a dog, a digit is a seven or a two, and a good model puts almost every example on the right side of the line. Single-cell perturbation data does not satisfy this assumption.

Two perturbations can be easy to tell apart as distributions yet impossible to tell apart cell by cell. Given enough cells, a statistical test will confidently report that their distributions differ, even when their responses overlap heavily in expression space. These are two different properties. Distinguishability improves as you collect more cells, because more of them land where only one perturbation has any mass. Separability does not improve with more data, because the region where the two distributions genuinely overlap does not shrink. So however distinguishable the perturbations are, accuracy measured one cell at a time is capped by that overlap.

We make this concrete on Tahoe-100M. A linear model, an MLP, and a Transformer all plateau at the same modest cell-level accuracy, with macro-F1 between 0.18 and 0.31. They do so despite millions of training cells and a 380-way problem that ought to reward capacity. The extra capacity does not help, because the bottleneck is the data and not the model. We then compare all pairs of perturbations as distributions, and almost every pair turns out to be distinguishable. So the signal is present. It does not show up in any single cell, only across the population.

The fix follows from the diagnosis. The signal is in the population, so we score the population instead of the cell. We train a standard single-cell classifier exactly as before and change only how we read it out at evaluation time. Instead of taking the argmax cell by cell, we average the classifier's output probabilities across all cells of a perturbation and take the argmax of that average. We call the resulting score the Classifier Discrimination Score (CDS). Nothing about training changes, and the cost stays linear in the number of cells. The same models now recover the true perturbation with rank-1 accuracy between 0.976 and 1.000. The classifier is scored on the same perturbations it was trained on, so this number shows the metric recognizes known perturbations reliably, not that perturbation prediction is solved.

CDS is one step away from the Perturbation Discrimination Score (PDS), the pseudobulk metric used in recent benchmarks \citep{wu2024perturbench, vinas2025systema, roohani2025virtual}. The two run the same averaging and differ only in the space they average in. PDS collapses each perturbation to a mean expression vector before measuring distances; CDS averages in the space of classifier probabilities, which the model has been trained to make discriminative. The space matters for how reliably each metric identifies the true perturbation from real cells, and PDS is the less reliable of the two. On the Virtual Cell Challenge, which scores predictions with PDS, CDS wins even with every cell in hand \citep{roohani2025virtual}. On Tahoe the gap opens up as cells grow scarce, the regime real experiments face. A ruler that misranks the ground truth will misrank the models measured against it, so a gap here translates directly into unreliable model comparisons.

\section{Background and related work}
\label{sec:related}

\paragraph{Metrics for perturbation responses.}
Most metrics fall into one of two categories. The cheap ones work on pseudobulk profiles, the mean
expression vector of a perturbation, and compare them with errors or correlations
\citep{squair2021confronting}. They are fast, but averaging throws away the cell-to-cell
variation that single-cell measurement was meant to capture. The expensive ones compare full
distributions. Maximum Mean Discrepancy \citep{gretton2012kernel} and energy distance
\citep{szekely2013energy, peidli2024scperturb} do this directly, but their cost grows with
the square of the number of cells, which is computationally prohibitive at atlas scale. The Perturbation
Discrimination Score tries to land in between \citep{wu2024perturbench, vinas2025systema, roohani2025virtual}. It ranks perturbations by the distance between their pseudobulk profiles,
so it stays cheap, but it inherits pseudobulk's blindness to within-perturbation structure.
Recent work shows that PDS is touchy in its own right, since its value moves with the choice
of distance and with a simple rescaling of the predictions \citep{liu2025effects}.

\paragraph{Classifiers as distribution comparisons.}
Using a classifier to compare distributions is an old idea. A classifier two-sample test
trains a model to tell two samples apart and reads its held-out accuracy as evidence that the
samples differ, an approach that has been used to judge how realistic a generative model's
samples are \citep{lopezpaz2017revisiting}. Learned kernels sharpen these tests and tie them
back to MMD \citep{liu2020learning}. What we do is the supervised, many-class version of the
same move. In place of a yes-or-no accuracy for a hypothesis test, we use the classifier's
averaged probabilities to pick which of many perturbations a population came from.

\paragraph{Population labels and overlap.}

Multiple-instance learning also turns a bag of cells into one label, but it solves a harder supervision problem. It knows the label of the bag and not of the individual cells, so it has to learn which cells matter, usually with attention \citep{ilse2018attention}. Our cells come labeled, so we train on them directly and pool only at the end. The overlap we rely on is not a new observation, and it has a biological reading. A perturbed population is rarely uniform: some cells respond and some look untouched, so the population is a mixture rather than a shifted copy of the control. Mixscape makes exactly this assumption in the simpler task of telling perturbed cells from controls, modeling the perturbed population as a blend of responders and non-responders \citep{papalexi2021characterizing}. That mixture structure is why a single cell can be uninformative while the population is not. It is the biological counterpart of the distinguishable-but-not-separable gap we exploit. We carry this observation into the many-perturbation setting and build a metric on it.

% Multiple-instance learning also turns a bag of cells into one label, but it solves a harder
% supervision problem. It knows the label of the bag and not of the individual cells, so it has
% to learn which cells matter, usually with attention \citep{ilse2018attention}. Our cells come
% labeled, so we train on them directly and pool only at the end. The overlap we rely on is not
% a new observation either. In the simpler task of telling perturbed cells from controls,
% Mixscape already models the perturbed population as a mix of cells that responded and cells
% that look untouched \citep{papalexi2021characterizing}. We take that observation to the
% many-perturbation setting and turn it into a recipe.

\section{Distinguishability is not separability}
\label{sec:gap}

We treat the cells of a perturbation $p$ as independent draws from a distribution
$\mathcal{D}_p$ over $\mathbb{R}^d$, where $d$ is the number of genes we keep. We will use two
terms repeatedly, so we define them now.

\begin{itemize}
\item Call $p$ and $q$ \textbf{distinguishable} if a two-sample test, run on cells from each,
rejects the idea that $\mathcal{D}_p$ and $\mathcal{D}_q$ are the same distribution. This
depends on how many cells you have. With enough of them, even a small real difference will
register \citep{lin2013research}.
\item Call $p$ and $q$ \textbf{separable} if you can look at a single cell and reliably say
which of the two it came from. This depends on how much the two distributions overlap in
space, and collecting more cells does nothing to change it.
\end{itemize}

In the benchmarks that taught us to trust accuracy, these two properties travel together.
Classes that differ also sit apart, so a model that can draw the boundary gets most cells
right. For single-cell perturbations, the two come apart.

\begin{figure}[htbp!]
\centering
\begin{subfigure}[b]{0.49\textwidth}
  \centering
  \includegraphics[width=\textwidth]{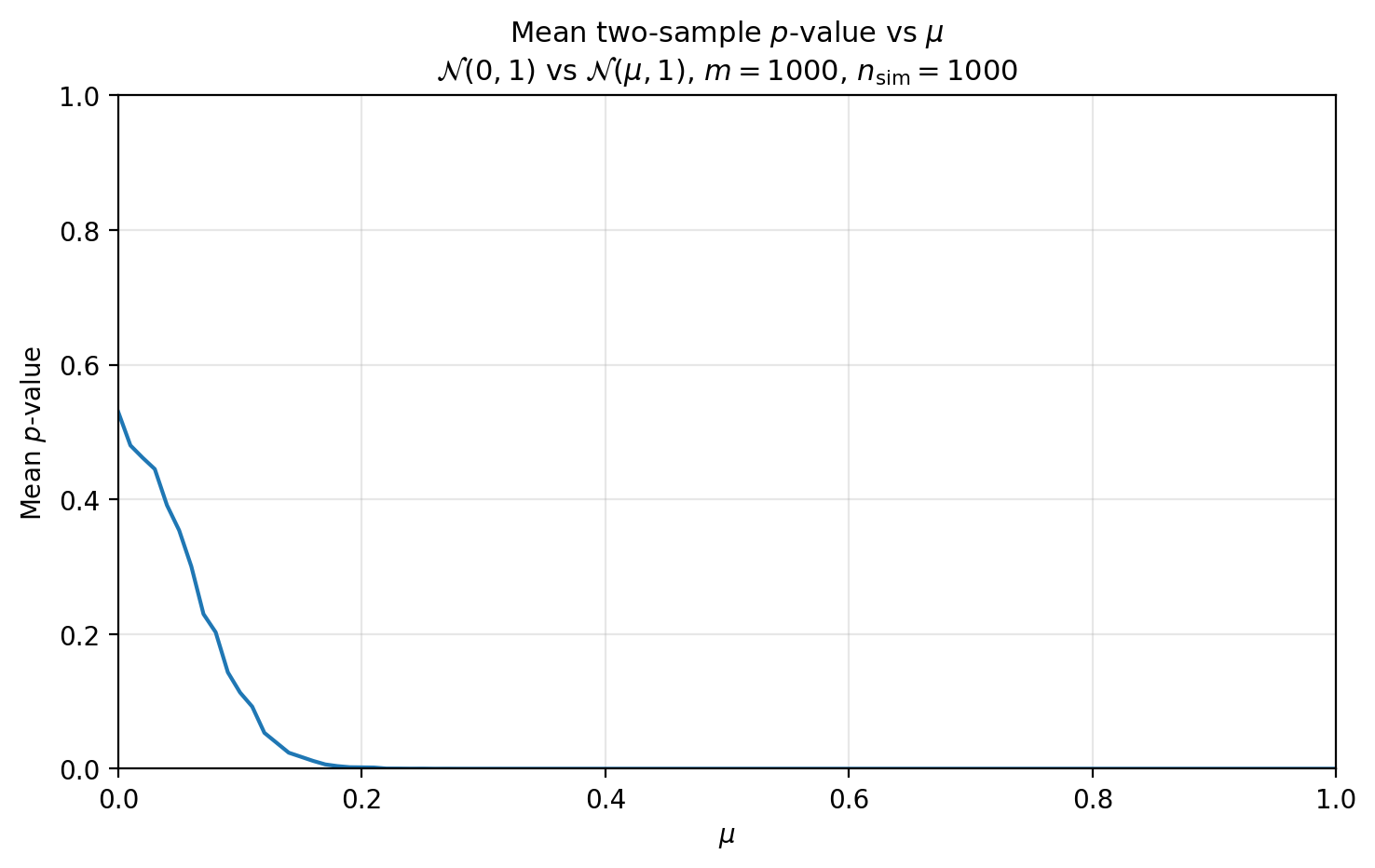}
\end{subfigure}
\hfill
\begin{subfigure}[b]{0.49\textwidth}
  \centering
  \includegraphics[width=\textwidth]{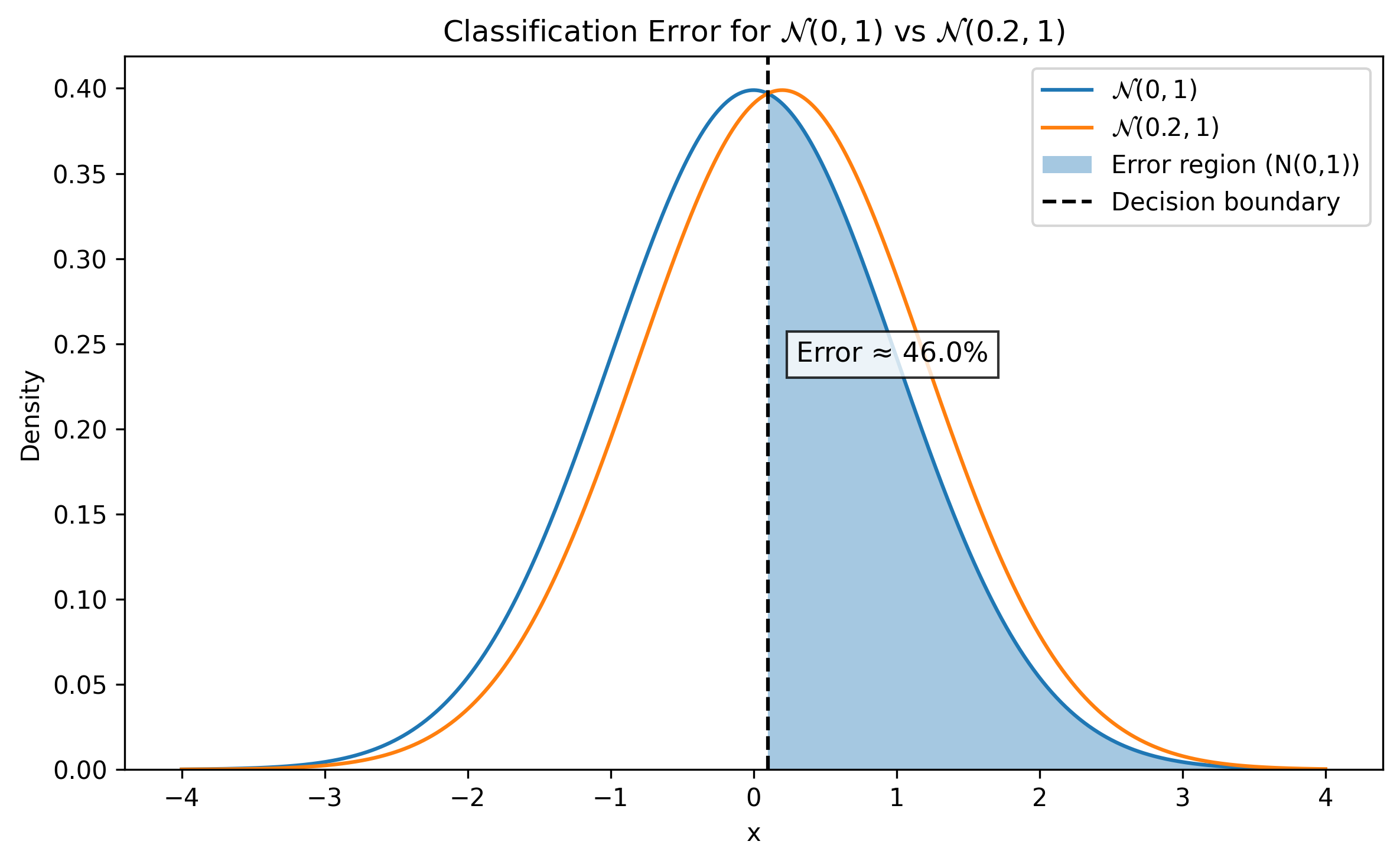}
\end{subfigure}
\caption{\textbf{Distinguishable but not separable.} Left: average two-sample $p$-value between
$\mathcal{N}(0,1)$ and $\mathcal{N}(\mu,1)$ as a function of $\mu$, with $1000$ samples per
group averaged over $1000$ simulations. The two distributions become statistically
distinguishable well before $\mu = 0.2$. Right: at $\mu = 0.2$ the densities overlap so heavily
that the best possible cell-level classifier still errs about $46\%$ of the time.
Distinguishability is a statistical property that grows with sample size. Separability is a
geometric property that does not.}
\label{fig:toy}
\end{figure}

\paragraph{A one-dimensional example.}
A simple example shows how far apart they can be. Let $\mathcal{D}_p = \mathcal{N}(0,1)$ and
$\mathcal{D}_q = \mathcal{N}(\mu, 1)$. Distinguishability grows quickly with $\mu$ and with
the number of samples. With a thousand cells per group, the average two-sample $p$-value is
already near zero by $\mu \approx 0.2$ (Figure~\ref{fig:toy}, left). Separability does not
move at all. At $\mu = 0.2$ the best possible classifier draws its boundary at $x = 0.1$ and
still gets the cell wrong almost half the time, since
$\mathbb{P}(X > 0.1) = 1 - \Phi(0.1) \approx 0.46$ (Figure~\ref{fig:toy}, right). The best
accuracy anyone can reach is about $54\%$, no matter how much data or how large the model. The
two distributions are clearly different, yet a single draw can rarely say which one it came
from.

This is only the one-dimensional picture. In high dimensions the overlap can occur in some
directions while the distributions stay apart in others, and the accounting is more involved.
Even so, distinguishable does not imply separable, so a classifier can hit a
hard floor on cell-level accuracy that comes from the data itself, even while the perturbations
stay perfectly distinguishable as distributions.

\section{A case study on Tahoe-100M}
\label{sec:realdata}

Single-cell perturbation data lives in exactly this regime, and Tahoe-100M lets us see both
sides of it. Our main testbed is the A549 cancer line from Tahoe-100M \citep{zhang2025tahoe}. We keep
to one cell line, so that differences between perturbations are not confounded by cell-line identity. We
work at two doses, $0.05$ and $5.00\,\mu$M. Each setting pairs $379$ chemical perturbations with
a control, for $380$ classes, over millions of cells. We vary the gene panel across
$d \in \{512, 1024, 2048, 4096\}$ highly variable genes. We split cells $70/15/15$ within each
perturbation, so every perturbation appears in every split (full dataset details in
Appendix~\ref{app:dataset}). The job is to recognize a perturbation we have already seen, from new cells, not to predict
one we have never met. A good metric should be able to tell, from real cells, which known
perturbation produced them.

We train three classifiers that differ mainly in capacity: a linear model, an MLP, and a
Transformer that treats each gene as a token \citep{vaswani2017attention}. All three use
class-weighted cross-entropy \citep{cui2019class} and are selected on validation macro-F1
(architecture and training details in Appendix~\ref{app:model}). The
first two numeric columns of Table~\ref{tab:tahoe} give their cell-level test scores. Accuracy
is low everywhere. More tellingly, it barely improves as the model grows. Macro-F1 goes
from $0.18$ to $0.22$ at the low dose and from $0.24$ to $0.31$ at the high dose as we move
from the linear model to the Transformer. A $380$-way problem with millions of training cells
should give a flexible model plenty to work with, yet the Transformer is barely ahead of a
linear baseline. When extra capacity buys almost nothing, the problem usually lies in the data
and not the model. This is the behavior the Gaussian example predicts when classes
overlap. As a metric, cell-level accuracy is uninformative here. It rates three very different
models within a few points of each other.

The plateau could mean two things. Maybe the perturbations are not actually different. Or
maybe they are different, but a single cell does not carry enough of that difference. We can
settle the first reading directly with a test on distributions. For two samples $X$ and $Y$,
the energy distance \citep{szekely2013energy} is
\[
\mathcal{E}(X,Y)=\frac{2}{nm}\sum_{i,j}\|x_i-y_j\|-\frac{1}{n^2}\sum_{i,i'}\|x_i-x_{i'}\|
-\frac{1}{m^2}\sum_{j,j'}\|y_j-y_{j'}\|,
\]
which is zero exactly when the two distributions match. A permutation test turns it into a
$p$-value by shuffling the pooled cells and seeing how often a random split looks as separated
as the real one. Across all of the roughly $72{,}000$ perturbation pairs at the high
dose, fewer than twenty fail to register a difference at $p < 0.05$ (see Appendix~\ref{app:distinguishability} for full details).
For all practical purposes, every pair of perturbations is distinguishable.

The two results point in opposite directions, but they do not conflict. The perturbations are
clearly different as distributions, while a classifier reading one cell at a time cannot
separate them, because the distributions overlap. The way out is to score the whole population
rather than one cell.

\section{The Classifier Discrimination Score}
\label{sec:aggregation}

The classifier from the previous section already holds the information we need, and the only
change is to read it out over a population rather than a single cell. Given a perturbation $p$
with evaluation cells $x_1, \dots, x_n$, the model maps each cell to a vector of class
probabilities $\pi_i = \mathrm{softmax}(f(x_i)) \in \Delta^{K-1}$. We average these vectors
across the population and take the top class,
\[
\bar{\pi}_p=\frac{1}{n}\sum_{i=1}^{n}\pi_i,
\qquad
\hat{p}=\arg\max_k\,[\bar\pi_p]_k.
\]
% The vector $\bar\pi_p$ is the Classifier Discrimination Score (CDS) of perturbation $p$. Its
% argmax gives the predicted perturbation, and its entries rank the candidate perturbations. Computing it reuses the trained classifier unchanged and takes a
% single pass over the cells, so its cost is linear in $n$.
We call this pooled vector $\bar\pi_p$ the CDS profile of the population, and the readout built on it the Classifier Discrimination Score (CDS). Its entries are indexed by the candidate perturbations, so the profile already ranks them: each entry scores how much the classifier attributes the population to that perturbation, and the top entry, $\arg\max_k [\bar\pi_p]_k$, is the predicted one. The readout reuses the trained classifier unchanged and takes a single pass over the cells, so its cost is linear in $n$.

Averaging is what turns a weak cell-level classifier into a reliable population-level one. Each
$\pi_i$ is noisy, and a cell drawn from the region where two perturbations overlap yields an
ambiguous prediction that no model can resolve from that cell alone. Across many cells of the
same perturbation these errors are unsystematic and largely cancel, while the consistent
population-level difference that the classifier learned from large training sets survives the
average. The pooled vector $\bar\pi_p$ therefore reflects the perturbation far more sharply
than any individual prediction.

This averaging also makes the relationship to PDS exact. Write a perturbation's summary as
$\phi(p) = \frac{1}{n}\sum_i \psi(x_i)$ for a feature map $\psi$. With $\psi$ the identity,
after subtracting a control, $\phi(p)$ is the pseudobulk profile that PDS compares. With
$\psi = \mathrm{softmax} \circ f$, $\phi(p)$ is the CDS vector. The two are the same average and
differ only in the space it is taken in, raw gene expression for PDS and a learned,
discriminative probability space for CDS \citep{liu2020learning}. In this sense CDS is the
many-class, identification form of a classifier two-sample test \citep{lopezpaz2017revisiting}.
Because the output is an average, CDS does not preserve the within-perturbation spread; it uses
that spread during training and reads it out through a map built to separate perturbations.

\paragraph{CDS recovers perturbation identity.}
Table~\ref{tab:tahoe} places cell-level accuracy next to the perturbation-level accuracy of CDS
for the same trained models. Cell-level accuracy stays between $0.18$ and $0.31$, while CDS runs
from $0.976$ to $1.000$. The capacity gap seen cell by cell closes, and the linear model, the
MLP, and the Transformer reach nearly the same accuracy once their predictions are pooled. The
model that scored $0.31$ per cell scores $0.997$ per perturbation, with nothing else changed.
What held the numbers down was the unit of evaluation, not the model. The same experiment on the Virtual Cell Challenge data follows the same pattern, which we report in Appendix~\ref{app:vcc}. 

\begin{table}[htbp!]
\centering
\caption{\textbf{The bottleneck is the unit of evaluation.} Tahoe-100M at both doses, best configuration per architecture. Cell-level accuracy and macro-F1 are modest and barely move with
capacity. Averaging the same model's cell-level probabilities across a perturbation's cells
(rightmost column) recovers near-perfect perturbation-level accuracy.}
\label{tab:tahoe}
\begin{tabular}{llccc}
\toprule
\textbf{Dose ($\mu$M)} & \textbf{Model} & \textbf{Cell-level accuracy} & \textbf{Cell-level macro-F1} & \textbf{Pert.-level accuracy} \\
\midrule
$0.05$ & Linear      & 0.185 & 0.182 & 0.976 \\
$0.05$ & MLP         & 0.205 & 0.211 & 0.995 \\
$0.05$ & Transformer & 0.218 & 0.215 & 0.990 \\
\midrule
$5.00$ & Linear      & 0.239 & 0.239 & 0.992 \\
$5.00$ & MLP         & 0.281 & 0.288 & 1.000 \\
$5.00$ & Transformer & 0.310 & 0.313 & 0.997 \\
\bottomrule
\end{tabular}
\end{table}

\section{CDS is a more reliable discriminator than PDS}
\label{sec:pds}

Forward models are ranked by scores like these, so a metric that misorders the true
perturbation will misorder the models built to predict it. Near-perfect accuracy on the full
test set is not, on its own, impressive: with thousands of cells in hand, even a raw pseudobulk
centroid will recognize a perturbation it has seen before. The sharper question is what happens
when cells run short, and whether averaging in a learned space buys anything over averaging raw
expression.

\paragraph{CDS and PDS as retrieval.}
We evaluate both scores the same way, as retrieval on the $70/15/15$ split. For each
perturbation we build a \emph{reference} from its training cells and a \emph{query} from its
held-out test cells, rank all reference perturbations against the query, and record the rank of
the true match. PDS works in gene-expression space. Its profile for a perturbation is the
differential pseudobulk $\Delta_p = \bar{x}_p - \bar{x}_{\mathrm{ctrl}}$, the mean expression of
the perturbation minus the mean of all control cells; the reference and query $\Delta_p$ are
formed from the training and test halves respectively, and references are ranked by their
Euclidean, Manhattan, or cosine distance to the query \citep{wu2024perturbench,
roohani2025virtual}. CDS works in the classifier's label space. Its reference is the trained
model and its query is the pooled profile $\bar\pi_p$ over the test cells; because the
classifier's outputs are indexed by the candidate perturbations, $\bar\pi_p$ already ranks them,
and the true match's rank is the position of its own label. Both build the reference from the
full training split and vary only the query. We shrink that query from $100\%$ down to $10\%$ of
a perturbation's test cells and report the rank-1 fraction,
the share of perturbations whose true match ranks first, together with the mean true-match rank.

\begin{figure}[htbp!]
\centering
\includegraphics[width=\textwidth]{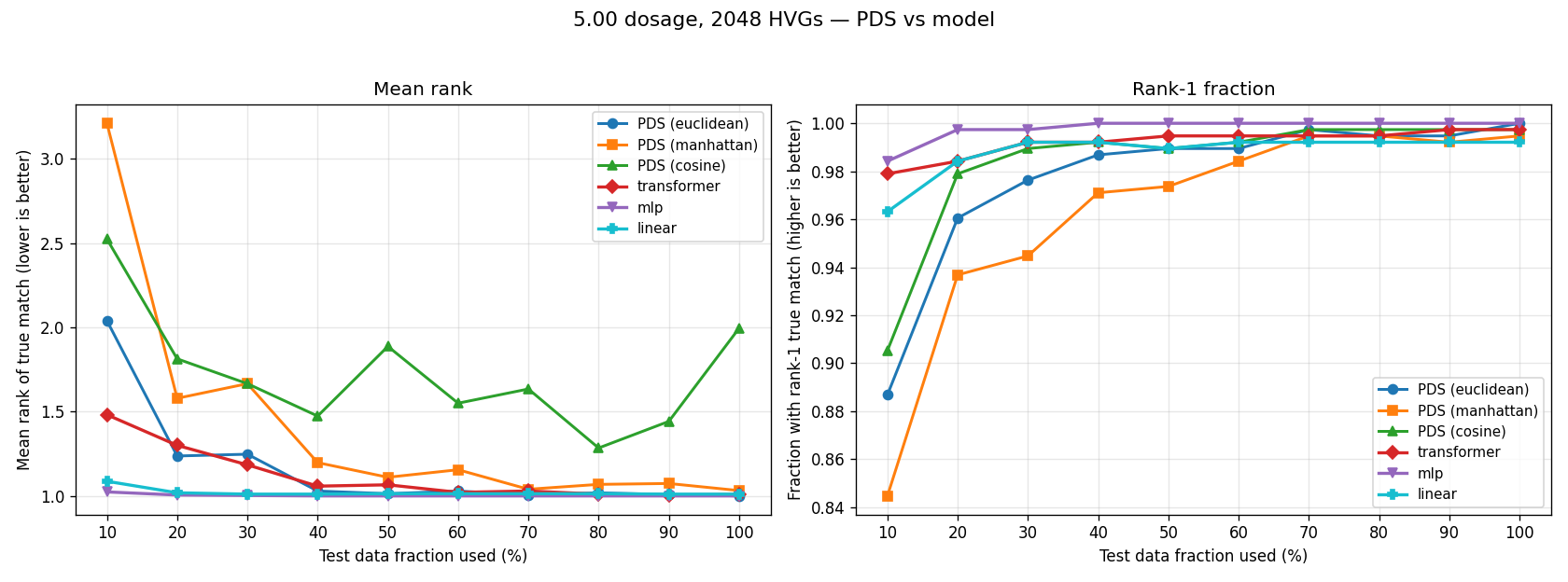}
\caption{Tahoe-100M at $5.00\,\mu$M:
rank-1 fraction against the fraction of evaluation cells used. CDS (linear, MLP, and
Transformer) stays near-perfect as cells become scarce, while the PDS variants drop, most of
all under the Manhattan distance. The axis starts at $0.84$, and every method sits above it.}
\label{fig:lowsample}
\end{figure}

\paragraph{Results on Tahoe-100M.}
CDS is far more sample-efficient (Figure~\ref{fig:lowsample}). The MLP and the Transformer stay near-perfect down to a tenth of the cells, while the PDS curves sag as cells
decrease. The drop is worst under the Manhattan distance and gentlest under cosine, which
lines up with the scale-sensitivity of $\ell_1/\ell_2$ PDS reported by \citet{liu2025effects}.
When only a few cells per perturbation are available, it pays to push the average through a
trained model rather than average raw expression. 

\begin{figure}[htbp!]
\centering
\includegraphics[width=\textwidth]{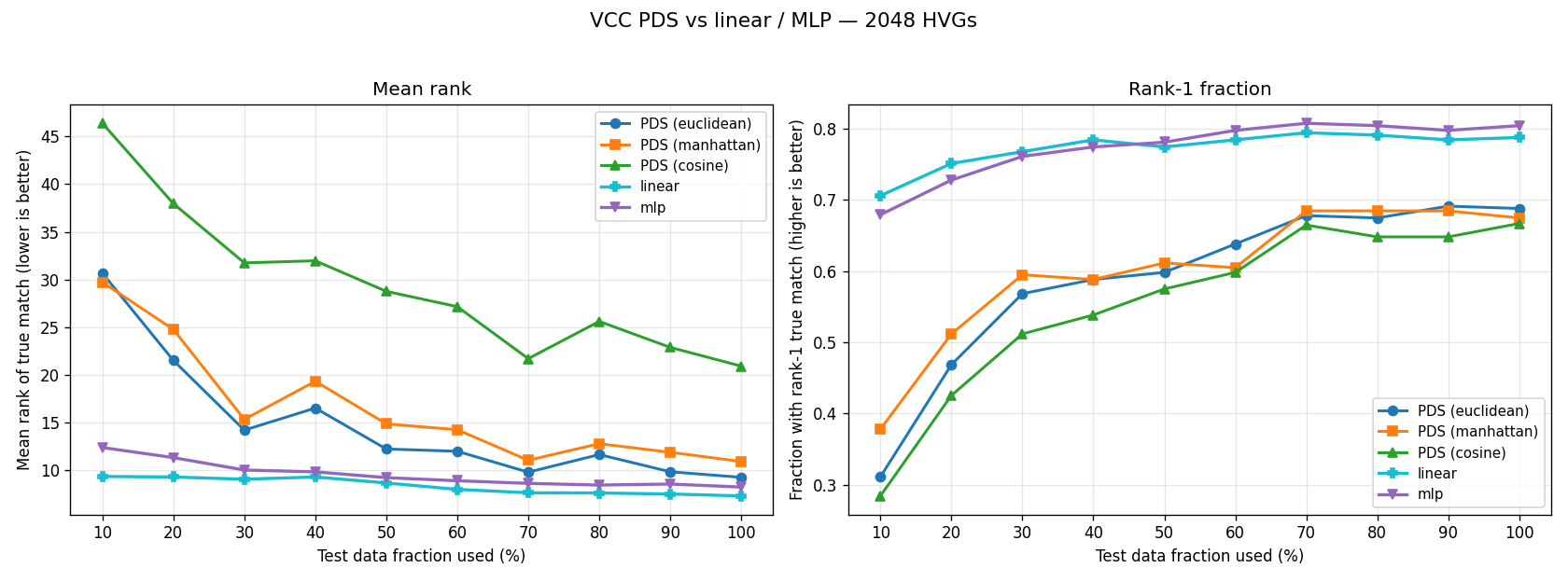}
\caption{VCC at $d = 2048$. Mean true-match rank (left, lower is better) and rank-1
fraction (right) against the fraction of evaluation cells used. CDS (linear and MLP) beats all
three PDS variants across fractions, and stays ahead even at full sample size.}
\label{fig:vcc}
\end{figure}

\paragraph{Results on the Virtual Cell Challenge dataset.}
A fair worry is that Tahoe is unusually generous, with millions of cells and a strong dose
signal. The Virtual Cell Challenge (VCC) data is a useful contrast \citep{roohani2025virtual}.
It is about an order of magnitude smaller, it uses genetic rather than chemical perturbations,
and it is measured in human embryonic stem cells instead of a cancer line, across $301$
classes. We build our own stratified splits, so these numbers should not be read against the
public challenge leaderboard (see Appendix~\ref{app:dataset}). Here the advantage no longer needs scarce cells to appear (Figure~\ref{fig:vcc}). With every
cell in hand, CDS reaches rank-1 of $0.86$ while the best PDS variant stalls near $0.70$, and
CDS beats all three PDS variants at every gene-panel size. The gap grows as the panel grows and
the underlying classifier gets stronger. Since VCC is a real benchmark that scores predictions
with PDS, this full-sample gap is direct evidence that PDS leaves usable discriminative signal
unmeasured. CDS does not reach its near-perfect Tahoe numbers here, which marks VCC as the
harder, weaker-signal case of the two.

\section{Discussion}
\label{sec:discussion}

Perturbation biology is moving toward large atlases and virtual-cell models trained on them. As those models improve, deciding which one is better depends on the metric used to compare them, which makes evaluation as consequential as modeling. This paper is about one part of that: how perturbation responses are scored. Cell-level accuracy conflates two questions. The first is whether perturbations differ at all, and on Tahoe they almost always do. The second is whether a single cell gives the difference away, and it usually does not, because the response distributions overlap. A metric built on single cells therefore reports the overlap in the data as much as the quality of the model. Ranked this way, the better model goes unrewarded: the score tracks how separable the data happens to be, not how good the model is.

CDS is the smallest change that fixes this. Training stays on single cells, but we move the decision to after a population-level average. It shares its averaging with PDS and its classifier with the two-sample test, and sits between them by averaging predictions rather than raw expression. It recovers the right perturbation from real cells more reliably than PDS: on Tahoe as cells grow scarce, and on VCC even at full sample.

% The one idea behind this paper is that perturbation effects in single-cell data live at the
% level of populations, not single cells, and that the we measure should match. Per-cell
% accuracy quietly mixes two questions that come apart in this setting. The first is whether
% perturbations differ at all, and they almost always do. The second is whether a single cell
% gives them away, and it usually does not. A score built on single cells therefore reports the
% overlap in the data as much as the quality of the model, which is why a Transformer barely
% edges out a linear model on Tahoe.

% Aggregation is the smallest change that fixes this. We keep training on single cells and move
% the decision to after a population-level average. Seen through the mean-embedding view, it is a
% close cousin of two methods at once. It shares its averaging with PDS and its classifier with
% the two-sample test, and it sits between them by averaging the classifier rather than the raw
% expression. Its edge over PDS is largest where evaluation cells are scarce, which is the
% regime that matters most in practice.

Two points are worth being plain about. First, CDS is scored on the perturbations its
classifier was trained to recognize, and that is by design. A discrimination metric only has to
know the perturbations in the benchmark, so near-perfect accuracy shows it recognizes them
reliably, not that we can predict unseen perturbations. Second, we measure the metric
on real cells whose labels are known, and we do not benchmark prediction models. That choice is
what makes the comparison clean. Which perturbation a population came from has a ground-truth
answer only because the cells are real and labeled, not generated by a model. Benchmarking
predictors is a separate question, in the spirit of recent metric analyses \citep{liu2025effects},
that we leave aside. What stays open is the floor and the reach of the metric. We do not estimate the Bayes error
directly. We read a cell-level floor off its fingerprint instead: the way accuracy refuses to
climb with capacity. And CDS recognizes only the perturbations in its training set, so a metric
that scores genuinely novel perturbations, or combinatorial ones where a cell carries several labels, is the natural next step.

\newpage

\small
\bibliographystyle{plainnat}
\bibliography{references}

\newpage

\appendix
\section{Datasets description}
\label{app:dataset}

We evaluate our methods on two large-scale single-cell perturbation datasets.
\paragraph{Tahoe-100M.}
Tahoe-100M \citep{zhang2025tahoe} is a large chemical perturbation atlas containing approximately 100 million single-cell transcriptomes collected across multiple human cancer cell lines, compounds, and doses. Throughout this work we restrict attention to the most represented cell line, the A549 cell line, which contains ${\sim}6.4\times10^{6}$ cells and consider two dosage settings, $0.05\,\mu$M and $5.00\,\mu$M, each together with the control condition. Both settings contain 380 perturbation labels (379 compounds plus control). We report results using subsets of highly variable genes (HVGs) of size $d \in \{512,1024,2048,4096\}$.

% Following the preprocessing provided with the dataset\todo{not sure what does it mean, we should also describe how we select highly variable genes}, we retain the top 5,000 highly variable genes and report results using subsets of size $d \in \{512,1024,2048,4096\}$\todo{lets report only the sizes actually used in the paper}.

\paragraph{Virtual Cell Challenge (VCC).}
The Virtual Cell Challenge dataset \citep{roohani2025virtual,roohanibehind} contains CRISPR-based genetic perturbations measured in human embryonic stem cells. After combining the official train, validation, and test partitions, the dataset contains 491,046 cells, 18,080 genes, and 301 perturbation labels (300 perturbations and one non-targeting control). Unlike Tahoe-100M, the dataset does not include multiple dosage levels. To keep the protocol aligned across datasets, we discard all-zero genes, then form new splits with the same rule used elsewhere in this project; for each \texttt{target\_gene}, cells are assigned 70\%/15\%/15\% to train/validation/test (stratified by cell, seed 42). HVGs are recomputed on \emph{training cells only} using Seurat v3\citep{stuart2019comprehensive} (raw counts, up to 4{,}096 genes), after which all splits are subset to the selected HVG panel and transformed with $\log(1+x)$.  We train the same linear and MLP classifiers under this preprocessing; the appendix additionally reports results for $d\in\{512,1024,2048,4096\}$.

\section{Models description}
\label{app:model}

We compare three supervised classifiers that predict perturbation identity from single-cell expression profiles. All models are trained with weighted cross-entropy (sklearn balanced class weights), optimized with Adam/AdamW, and selected by validation macro-F1 with early stopping (max 2{,}000 epochs). Unless noted otherwise, batch size is 128 and random seed is 42. Main Tahoe runs use $d=2048$ HVGs and learning rate $5\times10^{-5}$ with patience 30; VCC runs use patience 50 (linear at $10^{-4}$, MLP at $5\times10^{-5}$).

\paragraph{Linear.}
A single affine map $\mathbb{R}^{d}\rightarrow\mathbb{R}^{C}$, where $C$ is the number of perturbation classes (380 for Tahoe, 301 for VCC). This is the lowest-capacity baseline.

\paragraph{MLP.}
A fully connected network with ReLU activations and dropout ($p=0.2$) between hidden layers. For Tahoe we use four hidden layers of width 1{,}024; for VCC we use hidden sizes $(1024,512)$.

\paragraph{Transformer.}
Each gene is a token: a learned gene embedding plus a two-layer MLP projection of its expression value. Four class tokens are prepended and processed by a 4-layer Transformer encoder ($d_{\mathrm{model}}=256$, 4 heads, feedforward width 1{,}024, GELU, dropout 0 in our main runs). Token outputs are concatenated and passed to a linear classification head. This model is included as a high-capacity reference to test whether classifier performance is limited by expressivity rather than separability.

\section{Are perturbations distinguishable as distributions?}
\label{app:distinguishability}

We assess whether perturbations are distinguishable at the distribution level using pairwise two-sample testing based on the energy distance \citep{szekely2013energy}. For each pair of perturbations $p$ and $q$, we compare the corresponding empirical samples $X_p$ and $X_q$.

\subsection{Two-sample testing procedure}

We compute the energy distance $\mathcal{E}(X_p, X_q)$ between perturbation samples. Statistical significance is evaluated using a permutation test under the null hypothesis that $X_p$ and $X_q$ are drawn from the same distribution.

We form the pooled set $Z = X_p \cup X_q$ and repeatedly permute it into two subsets of sizes $|X_p|$ and $|X_q|$. For each permutation, we compute the energy distance, yielding a null distribution.

The p-value is defined as
\[
\text{p-value}(X_p, X_q)
= \mathbb{P}\big(\mathcal{E}(X'_p, X'_q) \geq \mathcal{E}(X_p, X_q)\big),
\]
where $(X'_p, X'_q)$ denotes a permutation-based split of $Z$.

\subsection{Hierarchical reordering}

To visualize global structure, we construct a hierarchical ordering of perturbations from the p-value matrix. We iteratively merge the pair of perturbations with the largest p-value (least distinguishable pair). After each merge, we recompute pairwise p-values from the merged sample sets.

This produces an ordering in which perturbations with higher distributional similarity are grouped earlier in the hierarchy.

\subsection{Global structure}

\begin{figure}[htbp!]
\centering
\includegraphics[width=\textwidth]{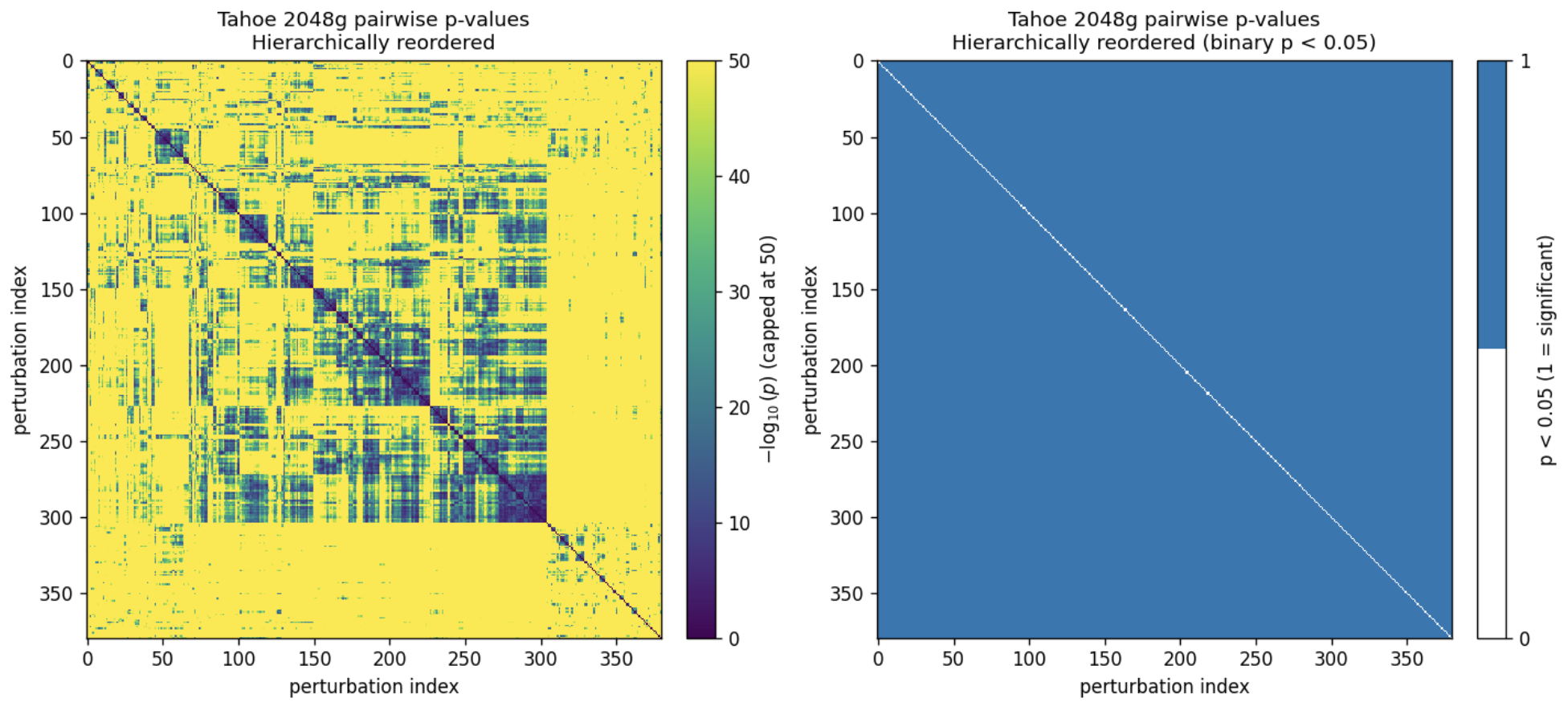}
\caption{\textbf{Almost everything is distinguishable.} Pairwise distribution-level $p$-values
on Tahoe-100M ($5.00\,\mu$M, $d = 2048$), with perturbations reordered by hierarchical
clustering. Left: $-\log_{10} p$, capped at $50$. Right: the same matrix thresholded at
$p < 0.05$, where black marks a rejected pair. Off the diagonal, fewer than twenty of about
$72{,}000$ pairs fail to reject. With thousands of cells per perturbation this is a low
statistical bar, as one would expect \citep{lin2013research}. Its only job here is to rule out
the trivial reading, that the classifier plateau reflects perturbations which are genuinely
identical.}
\label{fig:heatmap}
\end{figure}

Figure~\ref{fig:heatmap} shows the resulting pairwise p-value matrix for Tahoe-100M at dosage $5.00\,\mu$M and $d=2048$, reordered using the procedure above.

\section{Tahoe-100M dosage 0.05 results}

\begin{figure}[htbp!]
\centering
\includegraphics[width=\textwidth]{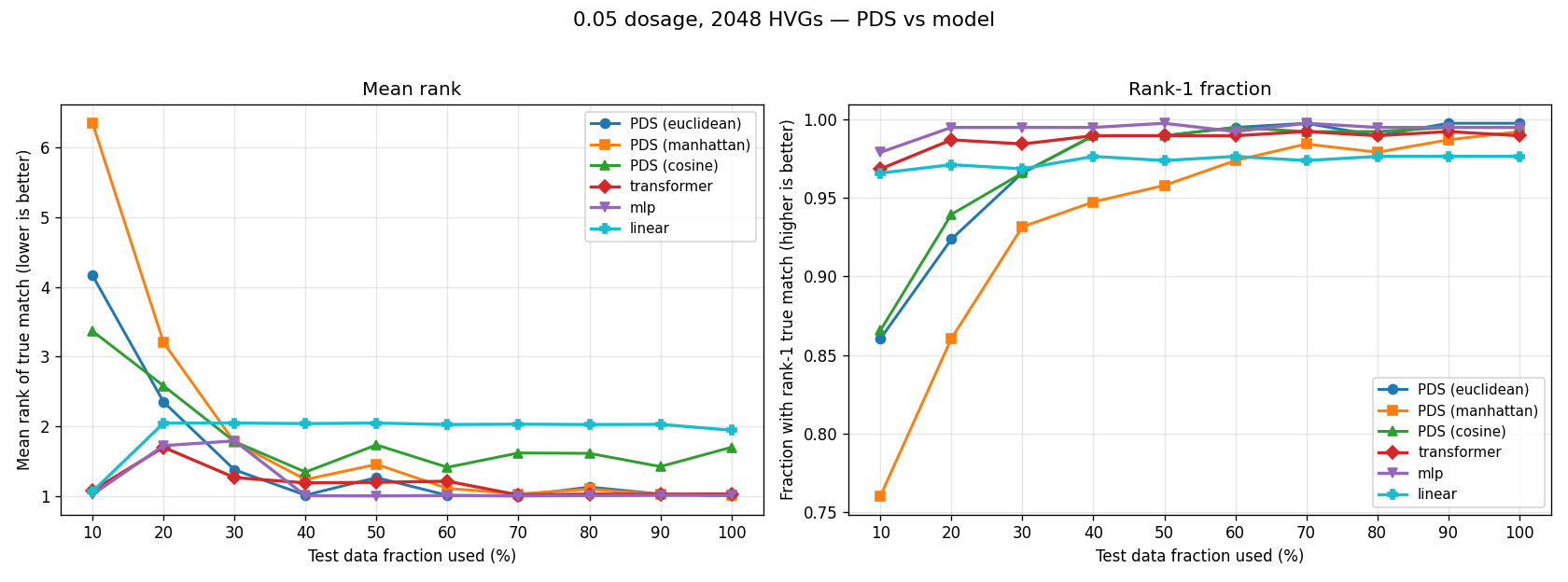}
\caption{Tahoe-100M at $0.05\,\mu$M:
rank-1 fraction against the fraction of evaluation cells used. CDS (linear, MLP, and
Transformer) stays near-perfect as cells become scarce, while the PDS variants drop, most of
all under the Manhattan distance. The axis starts at $0.75$, and every method sits above it.}
\label{fig:lowsample005}
\end{figure}

At $0.05\,\mu\mathrm{M}$, the same qualitative pattern holds as at $5.00\,\mu\mathrm{M}$: pairwise distributional tests reject almost all perturbation pairs, yet per-cell classifiers remain far from perfect. Under scarce evaluation cells, aggregation-based retrieval stays near-perfect while PDS distances degrade, especially under Manhattan distance (Figure~\ref{fig:lowsample005}).

\label{app:tahoe_005}
\newpage
\section{VCC datasaest results}
\label{app:vcc}

Table \ref{tab:vcc} reports CDS on the Virtual Cell Challenge dataset across gene-panel sizes. The pattern matches Tahoe-100M: cell-level accuracy and macro-F1 stay low and flat as model capacity grows, while pooling the same model's predictions into CDS lifts perturbation-level accuracy well above the cell-level numbers, though not to the near-perfect values seen on Tahoe-100M. As on Tahoe-100M, the advantage grows with the gene panel, since a larger panel gives the classifier more signal to pool.

\begin{table}[htbp!]
\centering
\caption{VCC results follow the same pattern as those of Tahoe-100M. Per-cell accuracy and macro-F1 seem to be capped by the data and do not benefit from increasing model capacity. While aggregation does not recover perfect accuracy like with Tahoe-100M we see that the results are much better}
\label{tab:vcc}
\begin{tabular}{llccc}
\toprule
\textbf{HVG } & \textbf{Model} & \textbf{Cell-level accuracy} & \textbf{Cell-level macro-F1} & \textbf{Pert.-level accuracy} \\
\midrule
$512$ & Linear      & 0.176 & 0.193 & 0.615 \\
$1024$ & Linear         & 0.246 & 0.295 & 0.704 \\
$2048$ & Linear        & 0.313 & 0.403 & 0.787 \\
$4096$ & Linear & 0.352 & 0.477 & 0.860 \\
\midrule
$512$ & MLP      & 0.170 & 0.207 & 0.621 \\
$1024$ & MLP         & 0.230 & 0.295 & 0.737 \\
$2048$ & MLP & 0.293 & 0.369 & 0.804 \\
$4096$ & MLP & 0.310 & 0.402 & 0.844 \\
\bottomrule
\end{tabular}
\end{table}

\end{document}